\documentclass[sigconf]{acmart}
\usepackage{fancyhdr}

\usepackage{booktabs}
\usepackage{graphicx}
\usepackage[utf8]{inputenc}
\usepackage[english]{babel}
\usepackage{amsthm}
\usepackage[switch]{lineno}
\newtheorem{definition}{Definition}
\newtheorem{proposition}{Proposition}

\usepackage{amsmath}
\usepackage{algorithm}
\usepackage{algorithmic}
\usepackage{graphicx}
\usepackage{multirow}
\usepackage{amsmath,mathtools}
\usepackage{color}

\renewcommand{\thefootnote}{\fnsymbol{footnote}}

\DeclareMathOperator*{\argmin}{arg\,min}

\AtBeginDocument{%
  \providecommand\BibTeX{{%
    \normalfont B\kern-0.5em{\scshape i\kern-0.25em b}\kern-0.8em\TeX}}}

\copyrightyear{2021} 
\acmYear{2021} 
\setcopyright{acmcopyright}
\acmConference[KDD '21]{Proceedings of the 27th ACM SIGKDD Conference on Knowledge Discovery and Data Mining}{August 14--18, 2021}{Virtual Event, Singapore}
\acmBooktitle{Proceedings of the 27th ACM SIGKDD Conference on Knowledge Discovery and Data Mining (KDD '21), August 14--18, 2021, Virtual Event, Singapore}
\acmPrice{15.00}
\acmDOI{10.1145/3447548.3467066}
\acmISBN{978-1-4503-8332-5/21/08}

\begin{document}
\fancyhead{}

\title{FIVES: Feature Interaction Via Edge Search for Large-Scale Tabular Data}


\author{
Yuexiang Xie$^{1,}\footnotemark[1]$,
Zhen Wang$^{1,}\footnotemark[1]$,
Yaliang Li$^{1,}\footnotemark[2]$,
Bolin Ding$^{1}$,\\
Nezihe Merve G{\"u}rel$^{2}$,
Ce Zhang$^{2}$,
Minlie Huang$^{3}$,
Wei Lin$^{1}$,
Jingren Zhou$^{1}$
}
\affiliation{
    \institution{
    \textsuperscript{\rm 1}Alibaba Group,
    \textsuperscript{\rm 2}ETH Z{\"u}rich,
    \textsuperscript{\rm 3}Tsinghua University}
    \country{}
}
\affiliation{
    \institution{\{yuexiang.xyx, jones.wz, yaliang.li, bolin.ding, weilin.lw, jingren.zhou\}@alibaba-inc.com}
    \country{}
}
\affiliation{
    \institution{\{nezihe.guerel, ce.zhang\}@inf.ethz.ch, aihuang@tsinghua.edu.cn}
    \country{}
}

\renewcommand{\authors}{Yuexiang Xie, Zhen Wang, Yaliang Li, Bolin Ding, Nezihe Merve G{\"u}rel, Ce Zhang, Minlie Huang, Wei Lin, Jingren Zhou}

\renewcommand{\thefootnote}{\fnsymbol{footnote}} 

\renewcommand{\shortauthors}{}

\begin{abstract}
High-order interactive features capture the correlation between different columns and thus are promising to enhance various learning tasks on ubiquitous tabular data.
To automate the generation of interactive features, existing works either explicitly traverse the feature space or implicitly express the interactions via intermediate activations of some designed models.
These two kinds of methods show that there is essentially a trade-off between feature interpretability and search efficiency.
To possess both of their merits, we propose a novel method named \textbf{F}eature \textbf{I}nteraction \textbf{V}ia \textbf{E}dge \textbf{S}earch (\textbf{FIVES}), which formulates the task of interactive feature generation as searching for edges on the defined feature graph.
Specifically, we first present our theoretical evidence that motivates us to search for useful interactive features with increasing order.
Then we instantiate this search strategy by optimizing both a dedicated graph neural network (GNN) and the adjacency tensor associated with the defined feature graph.
In this way, the proposed FIVES method simplifies the time-consuming traversal as a typical training course of GNN and enables explicit feature generation according to the learned adjacency tensor.
Experimental results on both benchmark and real-world datasets show the advantages of FIVES over several state-of-the-art methods.
Moreover, the interactive features identified by FIVES are deployed on the recommender system of Taobao, a worldwide leading e-commerce platform. Results of an online A/B testing further verify the effectiveness of the proposed method FIVES, and we further provide FIVES as AI utilities for the customers of Alibaba Cloud.
\end{abstract}

\ccsdesc[500]{Computing methodologies~Neural networks}
\ccsdesc[500]{Computing methodologies~Machine learning algorithms}

\keywords{Automated Machine Learning, Feature Interaction, Feature Graph}



\maketitle

\footnotetext[1]{Both authors contributed equally to this work.} 
\footnotetext[2]{Corresponding author.}
\renewcommand*{\thefootnote}{\arabic{footnote}}

\section{Introduction}
\label{sec:intro}
Data representation learning~\cite{bengio2013representation} is an important yet challenging task when applying machine learning techniques to real-world applications.
Recently, deep neural networks (DNN) have made impressive progress in representation learning for various types of data, including speech, image, and natural language, in which dedicated networks such as RNN~\cite{gers2002learning,cho2014learning}, CNN~\cite{krizhevsky2012imagenet} and Transformer~\cite{vaswani2017attention} are effective at learning compact representations to improve the predictive performance.
However, the tabular data, which are ubiquitous in many real-world applications such as click-through rate (CTR) prediction in recommendation and online advertising, are still heavily relied on manual feature engineering to explore higher-order feature space.

Specifically, tabular data are often stored as a SQL table where each instance corresponds to a row and each kind of feature corresponds to a column.
To achieve satisfactory performance on tabular data, feature interaction (i.e., feature crossing) is indispensable for industrial applications, which aims at generating high-order interactive features (i.e., combinatorial features, cross features) that are helpful for capturing the intrinsic correlation between different columns.
For example, a $3$-order interactive feature ``Gender $\otimes$ Age $\otimes$ Income'' can be discriminative for predicting the types of recommended commodities.
In practice, many data scientists search for useful interactive features in a trial-and-error manner, which has occupied a lot of their workloads.

Since such manual interactive feature generation is tedious and needs case-by-case experience that cannot be transferred easily, automatic feature generation~\cite{katz2016explorekit,kaul2017autolearn,shi2020safe,luo2020network}, one major topic of Automated Machine Learning (AutoML)~\cite{hutter2019automated} has attracted increasing attention from both academia and industry.
As a result, many kinds of DNNs~\cite{rendle2010factorization,juan2016field,shen2016deepcross,ijcai2017deep,lian2018xdeepfm,xu2018how,li2019fi,song2019autoint,su2021detecting} are deliberately designed and demonstrated their effectiveness in automatic feature generation.

Although these DNN-based methods can successfully express the interactive features by their hidden representations, the interpretability is weakened due to their black-box nature.
However, according to the requirements from both various scenarios of Alibaba and the customers of Alibaba Cloud, it is unfavorable to implicitly encode interactive features by a certain DNN, due to the following considerations:
(i)~\textbf{Overfitting issue}. The tabular data forms an extremely high-dimensional and sparse input, e.g., in the recommender system of Taobao, the dimension of raw features exceeds 2 billion with sparsity over $99.99\%$. With such an input space, DNNs can cause over-fitting issue easily with implicit interactive features.
(ii)~\textbf{Interpretability}. The hidden representation created by DNNs lacks of interpretability, which is required in many traditional industries (e.g., healthcare, manufacturing and finance). Even for the online services, product managers often do their analysis based on explicitly discovered interactive features.
(iii)~\textbf{Real-time requirement}. For industrial application scenarios, the real-time inference is always favored, and it becomes more important during a sale promotion (e.g., Double 11 promotion in Taobao) due to the tremendous traffic flow. In these and many more real-world scenarios, the inference cost of deep architecture can be prohibitive, while in contrast, explicitly generated interactive features can be fed into the lightweight models, with a negligible decrease in their inference speed.
This strong demand for explicitly identifying useful interactive features have also been discussed in previous works~\cite{kanter2015deep,katz2016explorekit,cheng2016wide,luo2019autocross,Tsang20feature}.

To meet the demand for explicitly generated interactive features, some of the related works, search-based methods~\cite{kanter2015deep,katz2016explorekit,luo2019autocross}, are proposed to traverse the high-order interactive feature space in a trial-and-error manner, with designed strategies for either pruning the search space or early stopping at each trial.
Although these search-based methods can satisfy the demand of explicit feature generation, when the scale of the considered tabular data is large, their required computing resources and training time can become intolerable.

To possess both feature interpretability and search efficiency, in this paper, we propose a novel method for automatic feature generation, called \textbf{F}eature \textbf{I}nteraction \textbf{V}ia \textbf{E}dge \textbf{S}earch (FIVES).
First, we propose to inductively search for an optimal collection of $(k+1)$-order features from the interactions between the generated $k$-order features and the original features.
A theoretical analysis (see Proposition~\ref{Proposition: UB of MI between feature product and output}) is provided to explain the intuition behind this search strategy---informative interaction features tend to come from the informative lower-order ones.
Then we instantiate this inductive search strategy by modeling the features as a \textit{feature graph} and expressing the interactions between nodes as propagating the graph signal via a designed graph neural network (GNN).
In the defined feature graph, the layer-wise adjacency matrix determines which original features should be selected to interact with which of the features generated by the previous layer.
In this way, we formulate the task of interactive feature generation as \textit{edge search}---learning the adjacency tensor.
Inspired by differentiable neural architecture search (NAS)~\cite{liu2018darts,noy2019asap}, we solve the edge search problem by alternatively updating the adjacency tensor and the predictive model, which transforms the trial-and-error procedure into a differentiable NAS problem. In particular, the learned adjacency tensor can explicitly indicate which interactive features are useful. 
We also parameterize the adjacency matrices in a recursive way where the dependencies are conceptually required by our search strategy. Such kinds of dependencies are ignored by most NAS methods~\cite{zhou2019bayesnas}.

To validate the effectiveness of the proposed method, we compare FIVES with both search-based and DNN-based methods, on benchmark datasets and real-world business datasets. The experimental results demonstrate the advantages of FIVES, as both an end-to-end predictive model and a feature generator. Further, the interactive features generated by FIVES are deployed in the recommender system of Taobao App, where online A/B testing during one week shows a significant improvement of CTR, indicating the usefulness of generated interactive features. Last but not the least, we provide FIVES as AI utilities to the customers of Alibaba Cloud.

In the following, we first present the details of the proposed method FIVES in Section~\ref{sec:methodology}, which consists of the inductively search strategy, the instantiation of search strategy via feature graph and a GNN-based method, optimization and interpretability of FIVES. Then, in Section~\ref{sec:exp}, we conduct quantitative evaluation to demonstrate the advantages of FIVES. We discuss related work in Section~\ref{sec:related}, and conclude the paper in Section~\ref{section:conlusion}.

\section{Methodology}
\label{sec:methodology}
\subsection{Preliminary}
\label{subsec:pre}
We consider the ubiquitous tabular data where each column represents a feature and each row represents a sample.
Without loss of generality, we assume that numeric features have been discretized and thus all considered features are categorical ones, for example, feature ``user city'' takes value from $\{\text{New York}, \text{London}, \ldots\}$, and feature ``user age'' takes value from $\{\text{under} \; 10, 10 \; \text{to} \; 20, \ldots\}$.
Based on these, we first define the high-order \textit{interactive feature}:
\begin{definition}
\label{interactivefeatures}
Given $m$ original features $F=[f_1,f_2,...,f_m]$, a $k$-order ($1\leq k \leq m$) interactive feature $f^k$ can be represented as the Cartesian product of $k$ distinct original features $f^k = f_{c_1} \otimes f_{c_2} \otimes ...\otimes f_{c_k}$
where each feature $f_{c_i}$ is selected from $F$.
\end{definition}
Since the interactive features bring in non-linearity, e.g., the Cartesian product of two binary features enables a linear model to predict the label of their XOR relation,
they are widely adopted to improve the performance of machine learning methods on tabular data. The goal of interactive feature generation is to find a set $\mathcal{F}=\{f_{1}^{k_1},f_{2}^{k_2},\ldots,f_{n}^{k_n}\}$ of such interactive features. 
To evaluate the usefulness of generated interactive features, we measure the performance improvements introduced by the generated features.
Following the majority of literature~\cite{liu2020autofis,luo2019autocross,li2019fi}, we take the CTR prediction as an application scenario where the usefulness can be measured by the AUC improvements.
The interactive feature generation task distinguishes from a standard classification task in its emphasis on identifying the useful interactive features that can be fed into other components of machine learning pipelines, including but not limited to other models, data preprocessing, result post-explanations. 

As the number of all possible interactions from $m$ original features is $O(2^{m})$, it is challenging to search for the optimal set of interactive features from such a huge space, not to mention the evaluation of a proposed feature can be costly and noisy.
Thus some recent methods model the search procedure as the optimization of designed DNNs, transforming it into a one-shot training at the cost of interpretability.
In the rest of this section, we present our proposed method---FIVES, which can provide explicit interaction results via an efficient differentiable search.

\subsection{Search Strategy}
\label{section:search_strategy}
As aforementioned, exhaustively traversal of the exponentially growing interactive feature space seems intractable.
By Definition~\ref{interactivefeatures}, any $k$-order interactive features could be regarded as the interaction (i.e., Cartesian product) of several lower-order features, with many choices of the decomposition.
This raises a question that could we solve the task of generating interactive features in a bottom-up manner?
To be specific, could we generate interactive features in an inductive manner, that is, searching for a group of informative $k$-order features from the interactions between original features and the group of $(k-1)$-order features identified in previous step?
We present the following proposition to provide theoretical evidence for discussing the question.
\begin{proposition}
\label{Proposition: UB of MI between feature product and output}
	Let $X_1,X_2$ and $Y$ be Bernoulli random variables with a joint conditional probability mass function, $p_{x_1,x_2\mid y} := \mathbb{P}(X_1 = x_1;X_2 = x_2\mid Y = y)$ such that $x_1, x_2, y \in \{0,1\}$. Suppose further that mutual information between $X_i$ and $Y$ satisfies $\mathcal{I} (X_i; Y) < L$ where $i \in \{1, 2\}$ and $L$ is a non-negative constant. If $X_1$ and $X_2$ are weakly correlated given $y \in \{0,1\}$, that is, $\Big\vert \frac{Cov(X_1,X_2\mid Y=y)}{\sigma_{X_1|Y=y} \sigma_{X_2|Y=y}}\Big\vert \leq \rho$, we have
	\begin{equation}
	\mathcal{I}(X_1X_2; Y ) < 2L + \log(2\rho^2+ 1).
	\end{equation}
\end{proposition}
We defer the proof to Appendix~\ref{proof}.
Specifically, the random variable $X$ and $Y$ stands for the feature and the label respectively, and the joint of $X$s stands for their interaction.
Recall that: 1) As the considered raw features are categorical, modeling each feature as a Bernoulli random variable would not sacrifice much generality;
2) In practice, the raw features are preprocessed to remove redundant ones, so the weak correlation assumption holds.
Based on these, our proposition indicates that, for small $\rho$ we have $\log(2\rho^2+ 1)\approx0$, thereby the information gain introduced by interaction of features is at most that of the individuals. This proposition therefore could be interpreted as---under practical assumptions, it is unlikely to construct an informative feature from the interaction of uninformative ones.
This proposition supports the bottom-up search strategy, as lower-order features that have not been identified as informative are less likely to be a useful building brick of high-order features.
Besides, the identified $(k-1)$-order features are recursively constructed from the identified ones of lower-orders, and thus they are likely to include sufficient information for generating informative $k$-order features.
We also empirically validate this strategy in Section~\ref{sec:vssearchbased} and Section~\ref{sec:usefulness}.
Although this inductive search strategy cannot guarantee to generate all useful interactive features, the generated interactive features in such a way are likely to be useful ones, based on the above proposition. 
This can be regarded as a trade-off between the usefulness of generated interactive features and the completeness of them, under the constraint of limited computation resources.

\subsection{Modeling}
\label{section:modelling}
To instantiate our inductive search strategy, we conceptually regard the original features as a \textit{feature graph} and model the interactions among features by a designed GNN.

First, we denote the feature graph as $G=(\mathcal{N},\mathcal{E})$ where each node $n_{i}\in \mathcal{N}$ corresponds to a feature $f_i$ and each edge $e_{i,j}\in \mathcal{E}$ indicates an interaction between node $n_i$ and node $n_j$.
We use $\mathbf{n}_{i}^{(0)}$ as the initial node representation for node $n_i$ that conventionally takes the embedding looked up by $f_i$ from the feature embedding matrix $\mathbf{W}_{F}$ as its value.
It is easy to show that, by applying a vanilla graph convolutional operator to such a graph, the output $\mathbf{n}_{i}^{(1)}$ is capable of expressing the $2$-order interactive features.
However, gradually propagating the node representations with only one adjacency matrix fails to express higher-order ($k > 2$) interactions.
Thus, to generate at highest $K$-order interactive features, we extend the feature graph by defining an adjacency tensor $\mathbf{A}\in\{0,1\}^{K\times m\times m}$ to indicate the interactions among features at each order, where each slice $\mathbf{A}^{(k)} \in \{0,1\}^{m\times m},k=0,\ldots,(K-1)$ represents a layer-wise adjacency matrix and $m=|\mathcal{N}|$ is the number of original features (nodes).
Once an entry $\mathbf{A}^{(k)}_{i,j},i,j \in {1,\ldots,m}$ is active, we intend to generate a $(k+1)$-order feature based on node $n_i$ by $\mathbf{n}_{i}^{(k-1)}\odot \mathbf{n}_{j}^{(0)}$ and synthesize these into $\mathbf{n}_{i}^{(k)}$.
Formally, with an adjacency tensor $\mathbf{A}$, our dedicated graph convolutional operator produces the node representations layer-by-layer, in the following way:
\begin{eqnarray}
\begin{split}
\mathbf{n}_{i}^{(k)}&=\mathbf{p}_{i}^{(k)}\odot\mathbf{n}_{i}^{(k-1)}\\
\text{where}\quad\mathbf{p}_{i}^{(k)}&=\text{MEAN}_{j|\mathbf{A}_{i,j}^{(k)}=1}\{\mathbf{W}_{j}\mathbf{n}_{j}^{(0)}\}.
\label{eq:convolution}
\end{split}
\end{eqnarray}
Here ``MEAN'' is adopted as the aggregator, and $\odot$ denotes the element-wise product. $\mathbf{W}_j$ is the transformation matrix for node $n_j$, and $n^{(0)}_i$ is the initial input to the GNN and described as the feature embeddings of node $n_i$.
Assume that the capacity of our GNN and embedding matrix is sufficient for $(\mathbf{W}_{j}\mathbf{n}_{j}^{(0)})\odot\mathbf{n}_{i}^{(0)}$ to express $f_{i}\otimes f_{j}$, we can show that the node representation at $k$-th layer $\mathbf{n}^{(k)}=[\mathbf{n}_{1}^{(k)},\ldots,\mathbf{n}_{m}^{(k)}]$ corresponds to the generated $(k+1)$-order interactive features: 
\begin{eqnarray}
\begin{split}
\mathbf{n}_{i}^{(1)}&=\text{MEAN}_{j|\mathbf{A}_{i,j}^{(1)}=1}\{\mathbf{W}_{j}\mathbf{n}_{j}^{(0)}\}\odot\mathbf{n}_{i}^{(0)}\\
&=\text{MEAN}_{j|\mathbf{A}_{i,j}^{(1)}=1}\{f_{j}\otimes f_{i}\},\\
\mathbf{n}_{i}^{(k)} &=\text{MEAN}_{j|\mathbf{A}_{i,j}^{(k)}=1}\{\mathbf{W}_{j}\mathbf{n}_{j}^{(0)}\}\odot\mathbf{n}_{i}^{(k-1)}\\
&\approx\text{MEAN}_{(c_{1},\ldots,c_{k})|\mathbf{A}_{i,c_j}^{(j)}=1,j=1,\ldots,k}\{f_{c_1}\otimes\cdots\otimes f_{c_k}\otimes f_{i}\}, \nonumber
\label{inductiveproof}
\end{split}
\end{eqnarray}
where the choices of which features should be combined are determined by the adjacency tensor $\mathbf{A}$.

As shown above, the feature graph and the associated GNN is capable of conceptually expressing our inductive search strategy. Thus from the perspective of feature graph, the task of generating interactive features is equivalent to learning an optimal adjacency tensor $\mathbf{A}$, so-called \textit{edge search} in our study.
In order to evaluate the quality of generated features, i.e., the learned adjacency tensor $\mathbf{A}$, we apply a linear output layer to the concatenation of node representations at each layer:
\begin{equation}
\hat{y}^{(k)}=\sigma(\mathbf{W}^{(k)}[\mathbf{n}_{1}^{(k)}:\cdots:\mathbf{n}_{m}^{(k)}]+b^{(k)}),
\end{equation}
where $\mathbf{W}^{(k)}$ and $b^{(k)}$ are the projection matrix and bias term respectively, and $\sigma(\cdot)$ denotes the sigmoid function.
We pack all the parameters as $\mathbf{\Theta}=(\mathbf{W}_{F},\mathbf{W}^{(k)},b^{(k)},\mathbf{W}_{j}|0\leq k<K,1\leq j\leq m)$.
Then we define a cross-entropy loss function (denoted as CE) for the joint of $\mathbf{A}$ and $\mathbf{\Theta}$:
\begin{equation}
\begin{gathered}
\mathcal{L}(\mathcal{D}|\mathbf{A},\mathbf{\Theta})=-\frac{1}{|\mathcal{D}|}\sum_{(x_i,y_i)\in \mathcal{D}}\frac{1}{K}\sum_{k=0}^{K-1}\text{CE}(y_i,\hat{y}_i^{(k)}),\\
\text{CE}(y_i,\hat{y}_i^{(k)}) = y_i\log(\hat{y}_i^{(k)})+ (1-y_i)\log(1-\hat{y}_i^{(k)}),
\label{eq:log_loss}
\end{gathered}
\end{equation}
where $\mathcal{D}$ is the considered dataset, $y_i$ denotes the ground truth label of $i$-th instance from $\mathcal{D}$, and $\hat{y}_i^{(k)}$ denotes its prediction based on the node representations from $k$-th layer.

Eventually, the edge search task could be formulated as a bilevel optimization problem:
\begin{equation}
\begin{gathered}
    \min_{\mathbf{A}}\mathcal{L}(\mathcal{D}_{\text{val}}|\mathbf{A},\Theta(\mathbf{A}))\\
    \text{s.t.}\qquad\Theta(\mathbf{A})=\argmin_{\mathbf{\Theta}}\mathcal{L}(\mathcal{D}_{\text{train}}|\mathbf{A},\mathbf{\Theta}),
\label{eq:bilevelopt}
\end{gathered}
\end{equation}
where $\mathcal{D}_{\text{train}}$ and $\mathcal{D}_{\text{val}}$ denote the training and the validation dataset respectively. 
Such a nested formulation in Eq.~\eqref{eq:bilevelopt} has also been studied recently in differentiable NAS~\cite{liu2018darts,cai2018proxylessnas,noy2019asap}, where the architecture parameters ($\mathbf{A}$ in our formulation) and network parameters ($\mathbf{\Theta}$ in our formulation) are alternatively updated during the search procedure. 
However, most of NAS methods ignored the dependencies among architecture parameters~\cite{zhou2019bayesnas}, which are critical for our task as the higher-order interactive features are generated based on the choice of previous layers.

\subsection{Differentiable Search}
\label{section:Optimization}
Directly solving the optimization problem in Eq.~\eqref{eq:bilevelopt} is intractable because of its bilevel nature and the binary values of $\mathbf{A}$.
Existing methods AutoInt~\cite{song2019autoint} and Fi-GNN~\cite{li2019fi} tackle the issue of binary $\mathbf{A}$ by calculating it on-the-fly, that is, the set of features to be aggregated for interaction is determined by a self-attention layer.
This solution enables efficient optimization, but the attention weights dynamically change from sample to sample. Thus it is hard to interpret these attention weights to know which interactive features should be generated.
On the other hand, there are some straightforward ways to learn a stationary adjacency tensor.
To be specific, we can regard $\mathbf{A}$ as Bernoulli random variables parameterized by $\mathbf{H}\in[0,1]^{K\times m\times m}$.
Then, in the forward phase, we sample each slice  $\mathbf{A}^{(k)}\sim\Pr(\cdot|\mathbf{H}^{(k)})$; and in the backward phase, we update $\mathbf{H}$ based on straight-through estimation (STE)~\cite{bengio2013estimating}.
This technique has also been adopted for solving problems like Eq.~\eqref{eq:bilevelopt} in some NAS studies~\cite{cai2018proxylessnas}.

Following the search strategy in Section \ref{section:search_strategy}, the adjacency tensor $\mathbf{A}$ should be determined slice-by-slice from $0$ to $(K-1)$.
In addition, since that which $k$-order features should be generated depend on those $(k-1)$-order features have been generated, the optimization of $\mathbf{A}^{(k)}$ should be conditioned on $\mathbf{A}^{(k-1)}$.
Our inductive search strategy would be precisely instantiated, only when such dependencies are modeled.
Thus, we parameterize the adjacency tensor $\mathbf{A}$ by $\mathbf{H}\in \mathbb{R}^{K\times m\times m}$ in this recursive way:
\begin{eqnarray}
\begin{split}
\mathbf{A}^{(k)}&\triangleq\varphi((\mathbf{D}^{(k-1)})^{-1}\mathbf{A}^{(k-1)} \sigma(\mathbf{H}^{(k)})),\\
\quad\mathbf{A}^{(0)}&\triangleq\mathbf{I}\text{ and }\mathbf{H}^{(0)}\triangleq\mathbf{I},
\label{eq:adjacency_matrix}
\end{split}
\end{eqnarray}
where $\varphi(\cdot)$ is a binarization function with a tunable threshold, and $\mathbf{D}^{(k-1)}$ is the degree matrix of $\mathbf{A}^{(k-1)}$ serving as a normalizer.

\begin{figure}[!h]
	\centering
	\includegraphics[width=0.42\textwidth]{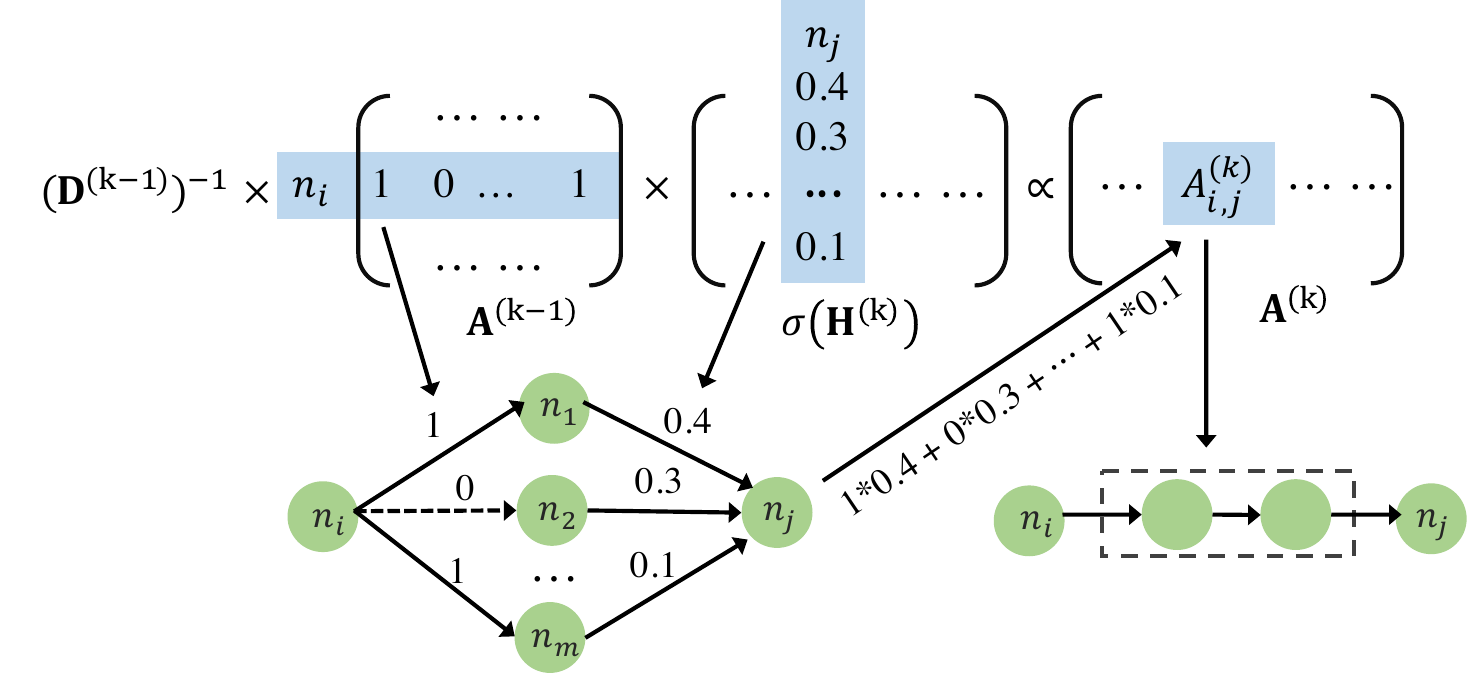}
	\vspace{-0.1in}
	\caption{The intuition behind Eq.~\eqref{eq:adjacency_matrix}.}
	\vspace{-0.2in}
	\label{figure:adj}
\end{figure}

We illustrate the intuition behind Eq.~\eqref{eq:adjacency_matrix} in Figure~\ref{figure:adj}.
Specifically, if we regard each $(k+1)$-order interactive feature, e.g., $f_{i}\otimes\cdots\otimes f_{j}$ as a $k$-hop path jumping from $n_i$ to $n_j$, then $\mathbf{A}^{(k)}$ can be treated as a binary sample drawn from the $k$-hop transition matrix where $\mathbf{A}_{i,j}^{(k)}$ indicates the $k$-hop visibility (or say accessibility) from $n_i$ to $n_j$.
Conventionally, the transition matrix of $k$-hop can be calculated by multiplying that of $(k-1)$-hop with the normalized adjacency matrix.
Following the motivation of defining an adjacency tensor such that the topological structures at different layers tend to vary from each other, we design $\sigma(\mathbf{H}^{(k)})$ as a layer-wise (unnormalized) transition matrix.
In this way, with Eq.~\eqref{eq:adjacency_matrix}, we make the adjacency matrix of each layer depend on that of previous layer, which exactly instantiates our search strategy.

Since there is a binarization function in Eq.~\eqref{eq:adjacency_matrix}, $\mathbf{H}$ cannot be directly optimized via a differentiable way w.r.t. the loss function in Eq.~\eqref{eq:log_loss}.
One possible solution is to use policy gradient~\cite{sutton2000policy}, Gumbel-max trick~\cite{jang2017categorical}, or other approximations~\cite{nayman2019xnas}.
However, it can be inefficient when the action space (here the possible interactions) is too large.

To make the optimization more efficient, we allow to use a soft $\mathbf{A}^{(k)}$ for propagation at the $k$-th layer, while the calculation of $\mathbf{A}^{(k)}$ still depends on a binarized $\mathbf{A}^{(k-1)}$:
\begin{eqnarray}
\begin{split}
\mathbf{A}^{(k)}&\triangleq(\mathbf{D}^{(k-1)})^{-1}\varphi(\mathbf{A}^{(k-1)})\sigma(\mathbf{H}^{(k)}),\\
\quad\mathbf{A}^{(0)}&\triangleq\mathbf{I}\text{ and }\mathbf{H}^{(0)}\triangleq\mathbf{I}.
\label{eq:new_aj_matrix}
\end{split}
\end{eqnarray}

However, the entries of an optimized $\mathbf{A}$ may still lie near the borderline (around $0.5$).
When we use these borderline values for generating interactive features, the gap between the binary decisions and the learned $\mathbf{A}$ cannot be neglected~\cite{noy2019asap}.
To fill this gap, we re-scale each entry of $\mathbf{A}^{(k)}$ through dividing it by a temperature $\tau$ before using it for propagation, which can be formatted as:
\begin{equation}
A^{(k)}_{i,j}\gets\sigma(\frac{\log [A^{(k)}_{i,j}/(1-A^{(k)}_{i,j})]}{\tau}),
\label{eq:gumbel}
\end{equation}
where $A^{(k)}_{i,j}$ denotes the entry of $\mathbf{A}^{(k)}$.
As $\tau$ anneals from $1$ to a small value, e.g., $0.02$ along the search phase, the re-scaled value becomes close to either $0$ or $1$. 

Finally, our modeling allows us to solve the optimization problem in Eq.~\eqref{eq:bilevelopt} with gradient descent method.
The whole optimization algorithm is summarized in Algorithm \ref{algorithm1}.
\begin{algorithm}[h]
	\caption{Optimization Algorithm for FIVES}
	\label{algorithm1}
	\begin{algorithmic}[1]
		\REQUIRE Feature graph $G=(\mathcal{N},\mathcal{E})$, highest order $K$, learning rate $\alpha_1, \alpha_2$, and \#epochs $T$
		\ENSURE Adjacency tensor $\mathbf{A}$, network parameters $\mathbf{\Theta}$ 
		\STATE Initialize $\mathbf{H}$ and $\mathbf{\Theta}$; and split data $\mathcal{D}$ into $\mathcal{D}_{\text{train}}$ and $\mathcal{D}_{\text{val}}$;
		\FOR {$t=1,2,\ldots,T$}
		\STATE Calculate $\mathbf{A}$ according to Eq.~\eqref{eq:new_aj_matrix};
		\STATE Propagate the graph signal for $K$ times according to Eq.~\eqref{eq:convolution};
		\STATE Update $\mathbf{\Theta}$ by descending $\alpha_1 \nabla_{\mathbf{\Theta}} \mathcal{L}(\mathcal{D}_{\text{train}}|\mathbf{A},\mathbf{\Theta})$;
		\STATE Update $\mathbf{H}$ by descending $\alpha_2 \nabla_{\mathbf{H}} \mathcal{L}(\mathcal{D}_{\text{val}}|\mathbf{A},\mathbf{\Theta})$;
		\ENDFOR
	\end{algorithmic}
\end{algorithm}

\subsection{Interpretability}
\label{sec:interpretability}
\begin{figure}[!ht]
	\centering
	\includegraphics[width=0.42\textwidth]{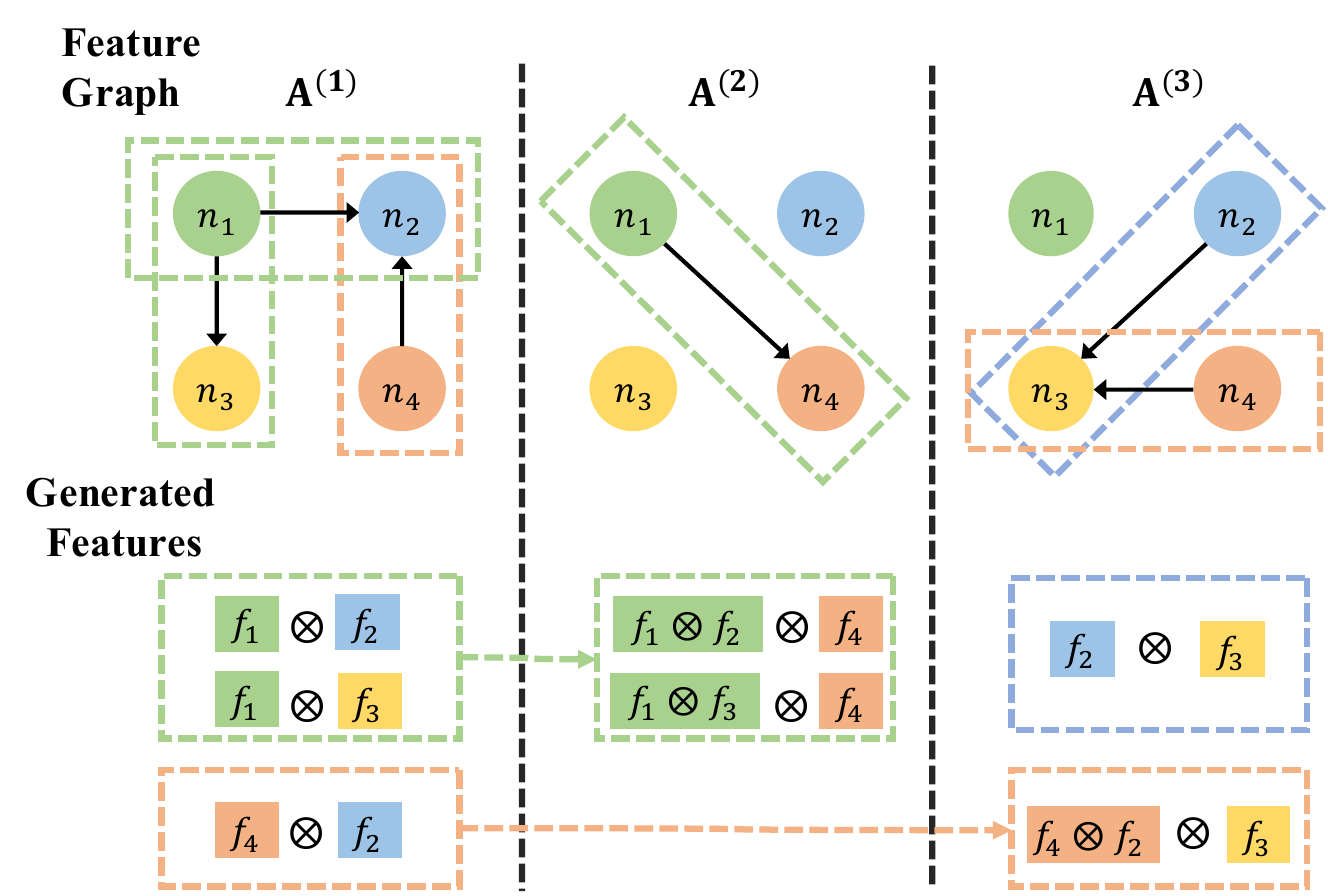}
	\vspace{-0.05in}
	\caption{Interactive feature derivation}
	\label{figure:reconstruct}
	\vspace{-0.1in}
\end{figure}

By Algorithm \ref{algorithm1}, the adjacency tensor $\mathbf{A}$ and the model parameters $\mathbf{\Theta}$ are learned and can solely serve as a predictive model.
Moreover, one can derive useful high-order interactive features according to the learned adjacency tensor $\mathbf{A}$, which means that FIVES can also serve as a feature generator.
In general, we are allowed to specify layer-wise thresholds for binarizing the learned $\mathbf{A}$ and then derive the useful $k$-order ($1\leq k\leq K$) interactive features suggested by $\mathbf{A}$ inductively.
An example of the interactive feature derivation is shown in Figure~\ref{figure:reconstruct}.

Specifically, given an adjacency matrix $\mathbf{A}^{(k)}$, which is the $k$-th slice of the binarized adjacency tensor $\mathbf{A}$, we can determine the feature graph (denoted as ``Feature Graph'' in the figure) at the $k$-th layer based on it, where
$\mathbf{A}^{(k)}_{i,j}=1$ represents there exists a directed edge from the start node $n_i$ to the end node $n_j$.
Then, we can derive useful interactive features (denoted as ``Generated Features'' in the figure) from the feature graphs layer-by-layer in an inductively manner.
At each layer, a directed edge means to apply the crossing operation (i.e., $\otimes$) between each of the features represented by the start node and that represented by the end node, which generates some interactive features.
As defined in Eq.~\eqref{eq:convolution}, the features represented by a node are different when it serves as a start node or an end node.
At the $k$-th layer, when a node $n_i$ serves as a start node, it represents the interactive features that have been generated by the edges starting from $n_i$ at the $(k-1)$-th order; elsewise, it just represents the corresponding original feature $f_i$.
Particularly, when $k=1$, no matter a node serves as a start node or an end node, it represents the corresponding original feature.

Take the instance in Figure~\ref{figure:reconstruct} as an example: In the middle column, $n_4$ is an end node in the connection $n_1 \rightarrow n_4$, thus it represents the original feature $f_4$; In the right column, $n_4$ is a start node in the connection $n_4 \rightarrow n_3$, thus it represents the interactive feature $f_4 \otimes f_2$, which is propagated from crossing in the previous order. 
The generated interactive features via crossing at each layer are synthesized in the start node and propagated to the next layer. The edge and propagation are denoted as the dotted frames and dotted arrows in different colors respectively in the figure.

Formally, the useful interactive features suggested by FIVES can be given as $\{f_{c_1}\otimes\cdots\otimes f_{c_k}\otimes f_i|\exists c_{1},\ldots,c_{k},\text{ s.t. }\mathbf{A}_{i,c_j}^{(j)}=1,j=1,\ldots,k$\}.
\section{Experiments}
\label{sec:exp}

We conduct a series of experiments to demonstrate the effectiveness of the proposed FIVES method, with the aims to answer the following questions. 
\textbf{Q1}: When the learned $(\mathbf{A}, \mathbf{\Theta})$ of FIVES solely serves as a predictive model, how it performs compared to state-of-the-art feature generation methods?  
\textbf{Q2}: Could we boost the performance of some lightweight models with the interactive features generated by FIVES? 
\textbf{Q3}: How do different components of FIVES contribute to its performance?
\textbf{Q4}: Are the interactions indicated by the learned adjacency tensor $\mathbf{A}$ really useful?
\textbf{Q5}: Can FIVES run as efficient as existing DNN-based methods?
\textbf{Q6}: Can we improve the CTR in a real-world e-commerce platform by deploying the interactive features generated by FIVES?

\textbf{Datasets}. We conduct experiments on five benchmark datasets that are widely adopted in related works, and we also include two more real-world business datasets. The datasets are randomly partitioned for a fair comparison, and their statistics are summarized in Table~\ref{table:statistics_of_datasets}. The availability of the benchmark datasets can be found in Appendix~\ref{datasets}.
\begin{table}[H]
	\centering
	\vspace{-0.1in}
	\caption{Statistics of datasets.}
	\vspace{-0.15in}
	\label{table:statistics_of_datasets}
	\begin{tabular}{lccc}
		\toprule
		 & \# Features & \# Train & \# Test \\
		\midrule
		Employee & 9 & 29,493 & 3,278 \\
		Bank & 20 & 27,459 & 13,729 \\
		Adult & 42 & 32,561 & 16,281 \\
		Credit & 16 & 100,000 & 50,000 \\
		Criteo & 39 & 41,256K & 4,584K \\
		Business1 & 53 & 1,572K & 673K \\
		Business2 & 59 & 25,078K & 12,537K \\
		\bottomrule
	\end{tabular}
	\vspace{-0.15in}
\end{table}

\begin{table*}[!ht]
	\centering
	\caption{Performance comparison for FIVES as a predictive model.}
	\vspace{-0.15in}
	\begin{tabular}{lccccccc}
		\toprule
		Method & Employee & Bank & Adult & Credit & Criteo & Business1 & Business2 \\
		\midrule
		LR & 0.8353 & 0.9377 & 0.8836 & 0.8262 & 0.7898 & 0.6912&  0.7121 \\
		DNN & 0.8510 & 0.9435 & 0.8869 & 0.8292 & 0.7779 & 0.6927 &0.6841\\
		FM & 0.8473 & 0.9434 & 0.8847  & 0.8278 & 0.7836 & 0.6888 & 0.7152\\
		Wide\&Deep & 0.8484 & 0.9416 & \textbf{0.8870} & 0.8299 & 0.7710 & 0.6941&  0.7128\\
		AutoInt & 0.8397  & 0.9393 & 0.8869 & 0.8301 & 0.7993 & 0.6960 & 0.7237\\
		Fi-GNN & 0.8260 & 0.9417 & 0.8813 & 0.8276 & 0.7964 & 0.6882 &0.7103 \\
		AutoFIS & 0.8454 & 0.9423 & 0.8840 & 0.8295 & 0.7916 & 0.6967 & 0.6981 \\
		\midrule
		FIVES & \textbf{0.8536} & \textbf{0.9446$^{\ast\ast}$} & 0.8863 & \textbf{0.8307$^{\ast}$} & \textbf{0.8006$^{\ast\ast}$} & \textbf{0.6984$^{\ast\ast}$} & \textbf{0.7276$^{\ast\ast}$}\\
		\bottomrule
	\end{tabular}
	\vspace{-0.05in}
	\label{table:main_result_AUC}
\end{table*}

\begin{table*}[!ht]
	\centering
	\caption{Performance improvement comparison for FIVES as a feature generator.}
	\vspace{-0.15in}
	\begin{tabular}{lccccccc|c}
		\toprule
		v.s. LR (base) & Employee & Bank & Adult & Credit & Criteo & Business1 & Business2 & Average\\
		\midrule
		Random+LR & -0.98\% &-0.04\%& -0.59\%& -0.04\%& -0.94\%&  0.15\%&  0.16\%& -0.33\%\\
		CMI+LR & 0.70\% & -0.07\% & -0.56\% & 0.02\% & -1.70\% & 0.30\% & 1.32\% & 0.01\%\\
		AutoCross+LR & 1.76\% & \textbf{0.16\%} &-0.65\% & \textbf{0.12\%} & 0.04\% & 0.04\% & 0.01\% & 0.21\%\\
		\midrule
		FIVES+LR & \textbf{1.79\%$^{\ast\ast}$} &0.01\% &\textbf{0.14\%$^{\ast\ast}$} &\textbf{0.12\%} &\textbf{0.26\%$^{\ast\ast}$} &\textbf{0.34\%$^{\ast}$} &\textbf{1.36\%}& \textbf{0.57\%} \\
		\bottomrule
	\end{tabular}
	\label{table:fg_result_AUC}
	\vspace{-0.1in}
\end{table*}

\textbf{Preprocessing}. We discretize numeric features into $\{10,100,1000\}$ equal-width buckets.
Then the numeric features are transformed into one-hot vector representations according to their bucket indices.
This follows the multi-granularity discretization proposed in~\cite{luo2019autocross}.
For all the rare feature category values (whose frequency is less than $5$), we assign them the same identifier.

\textbf{Metric}. Following existing works, we use AUC to evaluate the predictive performance.
A higher AUC indicates a better performance.
As has been pointed out in the previous studies~\cite{cheng2016wide,luo2019autocross,song2019autoint}, a small improvement (at \textbf{0.001-level}) in offline AUC evaluation can make a significant difference in real-world business predictive tasks such as CTR prediction in advertisements.

\subsection{FIVES as a predictive model (Q1)}
\label{sec:vsdnnbased}
As mentioned in Section \ref{section:Optimization}, the learned $(\mathbf{A},\mathbf{\Theta})$ of FIVES is a predictive model by itself.
We adopt the following methods as baselines, including those frequently used in practical recommender systems and state-of-the-art feature generation methods:
(1) LR: Logistic Regression with only  the original features (more settings of LR will be given in next part).
(2) DNN: The standard Deep Neural Network with fully connected cascade and a output layer with sigmoid function.
(3) FM~\cite{rendle2010factorization}: The factorization machine uses the inner product of two original features to express their interactions. 
(4) Wide\&Deep~\cite{cheng2016wide}: This method jointly trains wide linear models and deep neural networks.
(5) AutoInt~\cite{song2019autoint}: A DNN-based feature generation method, in which the multi-head self-attentive neural network with residual connections is proposed to model feature interactions.
(6) Fi-GNN~\cite{li2019fi}: It proposes to represent the multi-field features as a graph structure for the first time, and the interactions of features are modeled as the attentional edge weights. The implementation details of these methods can be found in Appendix~\ref{Implementation details of baselines}.

After hyperparameter optimization for all methods (see supplementary material for details), we use the optimal configuration to run each method for 10 times and conduct independent t-test between the results of FIVES and the strongest baseline method to show the significance: ``$\ast\ast$'' represents $p<0.01$ and ``$\ast$'' represents $p<0.05$. The experimental results are summarized in Table ~\ref{table:main_result_AUC}.

The experimental results demonstrate that FIVES can significantly improve the performance compared to baselines on most datasets. Especially on the large-scale datasets, FIVES outperforms all other baseline methods by a considerable large margin.

\begin{figure*}[!ht]
    \centering
    \includegraphics[width=0.9\textwidth]{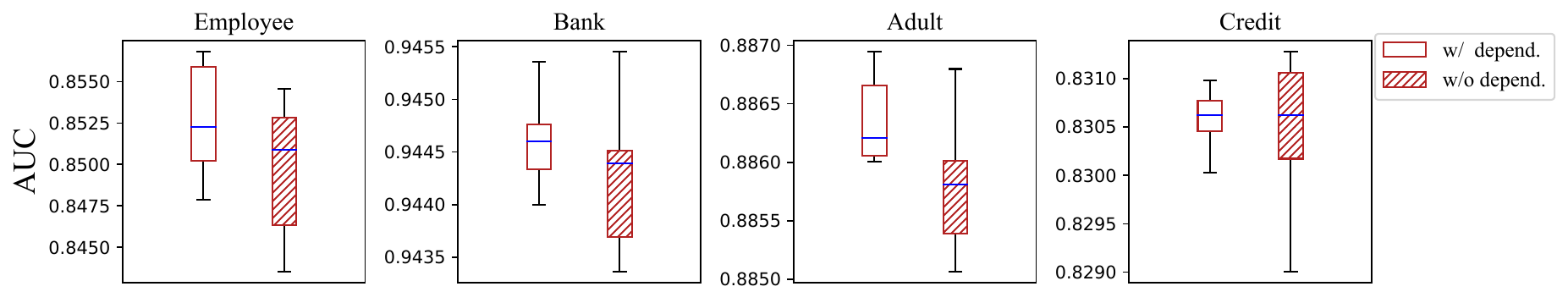}
    \vspace{-0.15in}
    \caption{Comparison of w/ and w/o modeling the dependencies between adjacency matrices.}
    \vspace{-0.15in}
    \label{figure:ablation_dependencies}
\end{figure*}

\begin{figure*}[!ht]
    \centering
    \includegraphics[width=0.9\textwidth]{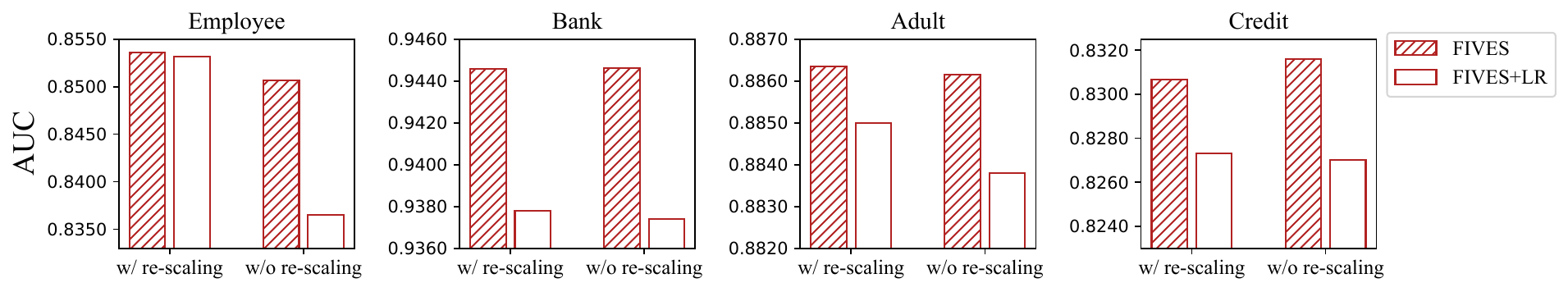}
    \vspace{-0.15in}
    \caption{Ablation study of the re-scaling applied to $\mathbf{A}$.}
    \label{figure:ablation_transformation}
    \vspace{-0.15in}
\end{figure*}

\subsection{FIVES as a feature generator (Q2)}
\label{sec:vssearchbased}
In practice, the generated interactive features are often used to augment original features, and then all of them are fed into some lightweight models to meet the requirement of inference speed. 
In specific, we adopt LR as the basic model to compare the usefulness of the interactive features discovered by different methods. It is worth pointing out that LR is widely adopted for such experiments in this area~\cite{luo2019autocross,rosales2012post,lian2018xdeepfm}, since it is simple, scalable, and interpretable.

As aforementioned, we can explicitly derive useful interactive features from the learned adjacency tensor $\mathbf{A}$.
We call this method FIVES+LR and compare it with the following methods:
(1) Random+LR: LR with original features and randomly selected interactive features;
(2) CMI+LR: conditional mutual information (CMI) as a filter to select useful interactive features from all possible $2$-order interactions;
(3) AutoCross+LR~\cite{luo2019autocross}: a recent search-based method, which performs beam search in a tree-structured space.
We also run the experiments for 10 times and analyze the results by t-test to draw statistically significant conclusions.
The results are summarized in Table \ref{table:fg_result_AUC}.

On all the datasets, the interactive features FIVES indicated consistently improve the performance of a LR.
The improvements are larger than that of other search-based method on most datasets, which is particularly significant on the largest one (i.e., Criteo).
The degeneration of some reference methods may be caused by redundant features that harden the optimization of a LR.

\subsection{Contributions of different components (Q3)}
\label{section:Q4}
\subsubsection{Modeling the dependencies between adjacency matrices}
As \cite{zhou2019bayesnas} pointed out that most existing NAS works ignore the dependencies between architecture parameters, which is indispensable to precisely express our inductive search strategy.
FIVES models such dependencies by recursively defining the adjacency matrices $\mathbf{A}^{(k)}$ (see Eq.~\eqref{eq:adjacency_matrix}).
In contrast,  
we define a variant of FIVES without considering such dependencies by parameterizing each $\mathbf{A}^{(k)}$ with its specific $\mathbf{H}^{(k)}$: $\mathbf{A}^{(k)}\triangleq\varphi( \sigma(\mathbf{H}^{(k)})),\quad\mathbf{A}^{(0)}\triangleq\mathbf{I}\text{ and }\mathbf{H}^{(0)}\triangleq\mathbf{I}$.
We conduct an ablation study to compare our recursive modeling with this independence modeling,
and the results in Figure \ref{figure:ablation_dependencies} clearly confirmed the effectiveness of the recursive modeling FIVES adopted.

\subsubsection{Filling the gap between differentiable search and hard decision}
\label{subsubsec:fillthegap}
Since a soft $\mathbf{A}$ is used in graph convolution (see Eq.~\eqref{eq:new_aj_matrix}), the learned $\mathbf{A}$ may lie near the borderline, which causes a gap between the binary decisions and the soft $\mathbf{A}$ used during search phase.
As \cite{noy2019asap} observed, such kind of gap hurts the performance of NAS.
To tackle this, we re-scale $\mathbf{A}$ with the temperature $\tau$ annealed.
Here is an ablation study to see the necessity of such a re-scaling.

From Figure~\ref{figure:ablation_transformation}, we observe that the re-scaling can noticeably reduce the performance gap between FIVES and FIVES+LR. Without this re-scaling, the interactions evaluated during search phase consist of softly weighted ones, which are inconsistent with the interactive features fed into LR.
By incorporating the re-scaling, almost all entries of $\mathbf{A}^{(k)}$ are forced to be binary before being taken into consideration by next layer, so that the decisions made at $(k+1)$-order are based on the exact $k$-order interactive features that will be fed into LR. In this way, the edge search procedure would be led by exact evaluation of what to be generated.

\subsection{Usefulness of generated interactive features (Q4)}
\label{sec:usefulness}

The above experiments evaluate the usefulness of the interactive features generated by FIVES from the perspective of augmenting the feature set for lightweight models.
Here, we further evaluate the usefulness of generated features from more different perspectives.

\begin{figure}[t]
	\centering
	\includegraphics[width=0.4\textwidth]{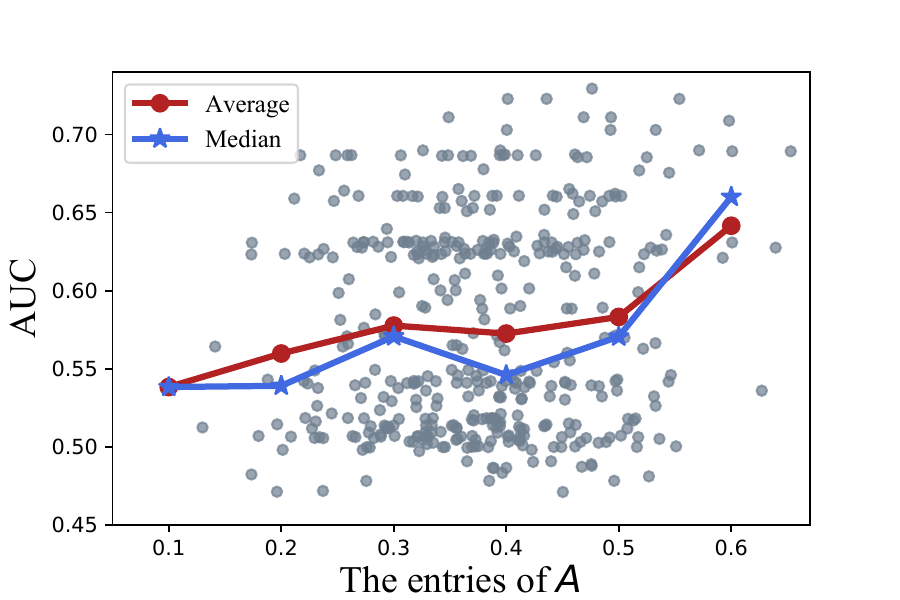}
	\vspace{-0.15in}
	\caption{Correlation between $\mathbf{A}_{i,j}^{(k)}$ and the AUC of the corresponding indicated feature.}
	\vspace{-0.15in}
	\label{figure:singlefeatureauc}
\end{figure}

First, we can directly calculate the AUCs of making predictions by each interactive feature.
Then we plot the features as points in Figure~\ref{figure:singlefeatureauc}, with the value of their entries $\mathbf{A}_{i,j}^{(k)}$ as x-axis and corresponding AUC values as y-axis, which illustrates a positive correlation between $\mathbf{A}_{i,j}^{(k)}$ and the feature's AUC.
This correlation confirms that the generated interactive features are indeed useful.

Another way to assess the usefulness of generated features is to compare different choices of the adjacency tensor $\mathbf{A}$.
From the view of NAS, $\mathbf{A}$ acts as the architecture parameters.
If it represents a useful architecture, we shall achieve satisfactory performance by tuning the model parameters $\mathbf{\Theta}$ w.r.t. it. 
Specifically, we can fix a well learned $\mathbf{A}$ and learn just the predictive model parameters $\mathbf{\Theta}$.
We consider the following different settings: (1) $\mathbf{\Theta}$ is learned from scratch (denoted as LFS); (2) $\mathbf{\Theta}$ is fine-tuned with the output of Algorithm~\ref{algorithm1} as its initialization (denoted as FT). We also show the performance of a random architecture (i.e., $\mathbf{A}$ is random initialized and will not be optimized) as the baseline (denoted as Random).

\begin{figure*}[!ht]
    \centering
    \includegraphics[width=0.9\textwidth]{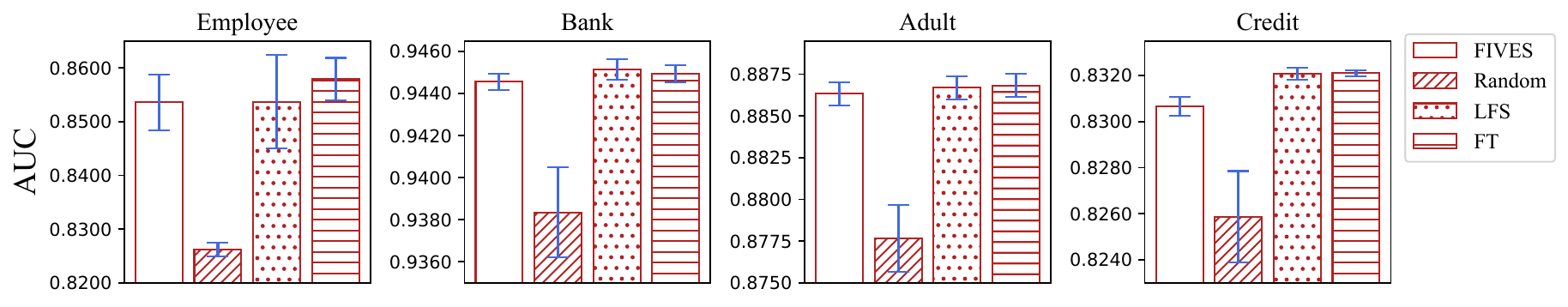}
    \vspace{-0.15in}
    \caption{Evaluation from the perspective of edge search.}
    \vspace{-0.1in}
    \label{figure:search_evaluation}
\end{figure*}

The experimental results in Figure~\ref{figure:search_evaluation} show the advantage of the searched architecture against a random one.
Since the architecture parameter $\mathbf{A}$ indicates which interactive features should be generated, the effectiveness of searched architecture confirms the usefulness of generated interactive features.
On the contrary, the randomly generated interactions can degrade the performance, especially when the original feature is relatively scarce, e.g., on Employee dataset.
Meanwhile, the improvement from ``FIVES'' to ``FT'' shows that, given a searched architecture (i.e., learned $\mathbf{A}$), improvement for predictive performance can be achieved via fine-tuning $\mathbf{\Theta}$.

\begin{figure*}[!ht]
	\centering
	\begin{minipage}[t]{0.42\textwidth}
		\centering
		\includegraphics[width=0.8\textwidth]{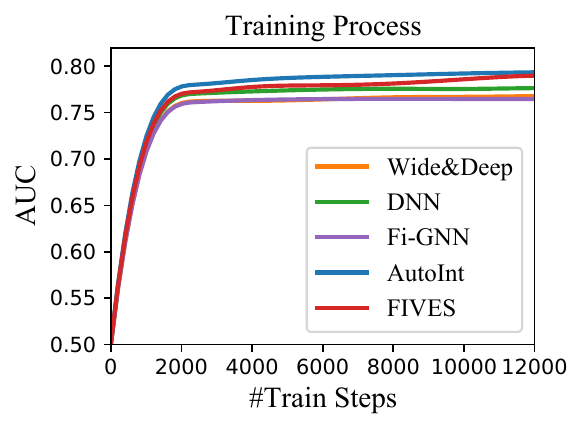}
	\end{minipage}
	\begin{minipage}[t]{0.42\textwidth}
		\centering
		\includegraphics[width=0.8\textwidth]{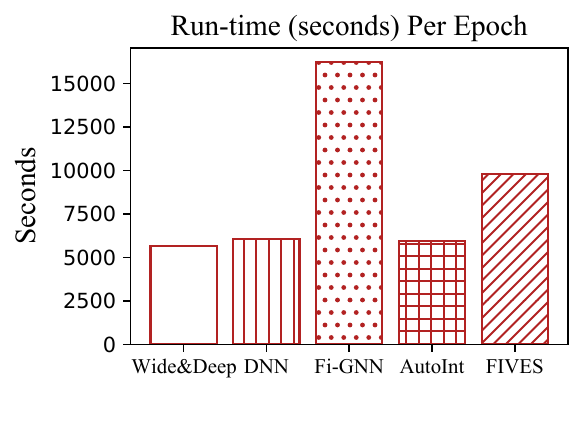}
	\end{minipage}
	\vspace{-0.15in}
	\caption{Efficiency comparisons between FIVES and some DNN-based methods on Criteo \label{figure:effciency_comparison}.}
	\vspace{-0.1in}
\end{figure*}

\subsection{Efficiency Comparisons (Q5)}
To study the efficiency of the proposed method, we empirically compare it against other DNN-based methods in terms of both convergence rate and run-time per epoch.
As Figure~\ref{figure:effciency_comparison} shows, although we formulate the edge search task as a bilevel optimization problem, FIVES achieves comparable validation AUC at each different number of training steps, which indicates a comparable convergence rate.
Meanwhile, the run-time per epoch of all the DNN-based methods is at the same order, even though, the run-time per epoch of both Fi-GNN and FIVES is relatively larger than that of others due to the complexity of graph convolutional operator.
Actually, with 4 Nvidia GTX1080Ti GPU cards, FIVES can complete its search phase on Criteo (traversal of 3 epochs) within $2$ hours.
In contrast, both dozens of hours and hundreds of times of computational resources are needed for search-based methods due to their trial-and-error nature.

\subsection{Online A/B Testing}
We conduct an online A/B testing for one week on a leading e-commerce platform, Tabobao, to verify the effectiveness of FIVES.
Specifically, we first train FIVES on the training data collected from search logs, and then derive the top 5 interactive features with the highest probability according to the learned adjacency tensor $\mathbf{A}$ (please refer to Section~\ref{sec:interpretability} for the derivation process).
The interactive features generated by FIVES are deployed on the online serving system, so that the deployed CTR prediction models (of various architectures) are fed with these features, which eventually affect the ranking of items.
The one-week observations show that the interactive features provided by FIVES contributes 0.42\% pCTR (page click-through rate, i.e., the ratio of page clicks and page views) and 0.28\% uCTR (user click-through rate, i.e., the ratio of user clicks and user views) improvements in the online A/B testing, which demonstrates the effectiveness of FIVES for real-world e-commerce platforms.

Moreover, we deploy FIVES as AI utilities on Alibaba Cloud\footnote{https://cn.aliyun.com/product/bigdata/product/learn} to empower a broader range of real-world businesses.
The data scientists from both Alibaba and the customers of Alibaba Cloud who want to generate useful interactive features can benefit from our proposed FIVES, especially those who demand for the interpretability of generated features.
\section{Related Works}
\label{sec:related}
To save data scientists from the tedious processes of manually feature generation, automatic feature interaction, one important topic of Automated Machine Learning (AutoML)~\cite{hutter2019automated,taking,zoller2019benchmark,khurana2018transformation}, has received a lot of attention from both academia and industry.
Recent works on automatic feature interaction can be roughly divided into two categories: search-based~\cite{kanter2015deep,katz2016explorekit,luo2019autocross,liu2020dnn2lr} and DNN-based~\cite{juan2016field,shen2016deepcross,ijcai2017deep,lian2018xdeepfm,xu2018how,rendle2010factorization,li2019fi,song2019autoint,wang2020dcn,wang2017deep}.

The search-based methods focus on designing different search strategies that prune as much of the candidates to be evaluated as possible, while aiming to keep the most useful interactive features.
For example, ExploreKit~\cite{katz2016explorekit} evaluates only the top-ranked candidates where the feature ranker has been pre-trained on other datasets.
AutoCross~\cite{luo2019autocross} incrementally selects the optimal candidate from the pairwise interactions of current features.
Although these mechanisms can trim the search space to be traversed, due to their trial-and-error nature, the needed time and computing resources are usually intolerable in practice.

On the other hand, the DNN-based methods design specific neural architectures to express the interactions among different features.
For example, AutoInt~\cite{song2019autoint} and Fi-GNN~\cite{li2019fi} exploit the self-attention mechanism~\cite{vaswani2017attention} for weighting features in their synthesis, and achieve leading performances with just a one-shot training course. 
But this advantage comes at the cost of implicit feature interactions as it is hard to exactly interpret which interactive features are useful from the attention weights.
Actually, it is in great demand that useful interactive features can be explicitly expressed, since they can be incorporated to train lightweight predictive models to satisfy the requirement of real-time inference. 
Some recent works~\cite{song2020towards,liu2020autofis} explicitly parameterize the relationships among features and optimize in a differentiable way, but without a well-designed search strategy they still sacrifice either the interpretability of search-based methods or the efficiency of DNN-based methods.
\section{Conclusions}
\label{section:conlusion}
Motivated by our theoretical analysis,
we propose FIVES, an automatic feature generation method where an adjacency tensor is designed to indicate which feature interactions should be made.
The usefulness of indicated interactive features is confirmed from different perspectives in our empirical studies.
FIVES provides such interpretability by solving for the optimal adjacency tensor in a differentiable manner, which is much more efficient than search-based method, while also preserving the efficiency benefit of DNN-based method.
Extensive experiments and an one-week online A/B testing show the advantages of FIVES as both a predictive model and a feature generator.

\section*{Acknowledgment}
This work was supported by Alibaba Group through a donation to the ETH Foundation.

\bibliographystyle{kdd}
\bibliography{kdd}

\appendix
\section{Proof of Proposition 1} 
\label{proof}

We first re-state the Proposition 1 in the paper here.
\begin{proposition}
Let $X_1,X_2$ and $Y$ be Bernoulli random variables with a joint conditional probability mass function, $p_{x_1,x_2\mid y} := \mathbb{P}(X_1 = x_1;X_2 = x_2\mid Y = y)$ such that $x_1, x_2, y \in \{0,1\}$. Suppose further that mutual information between $X_i$ and $Y$ satisfies $\mathcal{I} (X_i; Y) < L$ where $i \in \{1, 2\}$ and $L$ is a non-negative constant. If $X_1$ and $X_2$ are weakly correlated given $y \in \{0,1\}$, that is, $\Big\vert \frac{Cov(X_1,X_2\mid Y=y)}{\sigma_{X_1|Y=y} \sigma_{X_2|Y=y}}\Big\vert \leq \rho$, we have
\begin{equation}
	\mathcal{I}(X_1X_2; Y ) < 2L + \log(2\rho^2+ 1).
\end{equation}
\end{proposition}

\begin{proof}
As explained in Section 2.2 of the paper, we use $X_{1}X_{2}$ to denote the interaction $f_{1}\otimes f_{2}$.
By Definition 1, the interaction is defined to be the Cartesian product of the individual features.
In this sense, $X_{1}X_{2}$ could be regarded as a random variable constructed by some bijective mapping from the tuple $(X_1, X_2)$.
In our method, the interaction is expressed via the graph convolutional operator.
Although we have assumed such a modeling to be expressive enough, for rigorous analysis, we'd better regard $X_{1}X_{2}$ as a non-injective mapping from $(X_1, X_2) \in \{0, 1\}\times\{0, 1\}$ to $X_1X_2 \in \{0, 1\}$, and thus we have $\mathcal{H}(Y|X_1X_2) \geq \mathcal{H}(Y|X_1, X_2)$.
Therefore:
\begin{align}
\label{Eq: product UB}
\mathcal{I}(X_1X_2; Y) &=\mathcal{H}(Y)-\mathcal{H}(Y|X_1X_2)\nonumber\\
&\leq \mathcal{H}(Y)-\mathcal{H}(Y|X_1, X_2) = \mathcal{I}(X_1, X_2; Y)\nonumber\\
&\leq \mathcal{H}(X_1)+\mathcal{H}(X_2)-\mathcal{H}(X_1|Y)\nonumber\\
&\quad -\mathcal{H}(X_2|Y)+{1}/{d},
\end{align}
where $d$ is the so called incremental entropy:
\begin{equation}
    d={1}/\big({\mathcal{H}(X_1|Y)+\mathcal{H}(X_2|Y)-\mathcal{H}(X_1, X_2|Y)}\big).
\end{equation}
Note that $\mathcal{I}(X_1; Y)=\mathcal{H}(X_1)-\mathcal{H}(X_1|Y)< L$ and $\mathcal{I}(X_2; Y)=\mathcal{H}(X_2)-\mathcal{H}(X_2|Y)< L$, we further have
\begin{equation}\label{Eq: proposition}
\begin{split}
    \mathcal{I}(X_1, X_2; Y) &= \mathcal{I}(X_1; Y)+\mathcal{I}(X_1; Y)+{1}/{d}\\
    &< 2L+{1}/{d}.
\end{split}
\end{equation}
To prove the Proposition, it remains to prove that:
\begin{equation}
    1/d \leq \log(2\rho^2+1).
\end{equation}

We start with deriving the mathematical expression for $\frac{Cov(X_1, X_2|Y=y)}{\sigma_{X_1|Y=y}\sigma_{X_2|Y=y}}$ in terms of $p_{x_1, x_2|y}$, $x_1, x_2, y \in \{0, 1\}$. For convenience, we define the following notations:
\begin{equation}
    \begin{split}
        A = p_{0, 0|y}, \quad B = p_{0, 1|y},\\
        C = p_{1, 0|y}, \quad D = p_{1, 1|y}.
    \end{split}
\end{equation}
Note the following mathematical relations:
\begin{equation}
    \begin{split}
    &\mathbb{P}(X_1=0|Y=y)=p_{0, 0|y}+p_{0, 1|y} = A+B, \\ 
    &\mathbb{P}(X_1=1|Y=y)=p_{1, 0|y}+p_{1, 1|y} = C+D,\\
    &\mathbb{P}(X_2=0|Y=y)=p_{0, 0|y}+p_{1, 0|y} = A+C,\\ 
    &\mathbb{P}(X_2=1|Y=y)=p_{0, 1|y}+p_{1, 1|y} = B+D.
    \end{split}
\end{equation}
Using the above relations, it is straightforward to show that
\begin{equation}\label{Eq: std's}
    \begin{split}
    \sigma_{X_1|Y=y}&=\sqrt{(p_{1, 0|y}+p_{1, 1|y})(1-p_{1, 0|y}-p_{1, 1|y})}\\
    &=\sqrt{(C+D)(A+B)}\\
    \sigma_{X_2|Y=y}&=\sqrt{(p_{0, 0|y}+p_{1, 0|y})(1-p_{0, 0|y}-p_{1, 0|y})}\\
    &=\sqrt{(A+C)(B+D)}.
    \end{split}
\end{equation}
Next, we derive the expression for $Cov(X_1, X_2|Y=y)$:
\begin{equation}\label{Eq: cov}
    \begin{split}
     & \quad Cov(X_1, X_2|Y=y)\\
     &=p_{0, 0|y}p_{1, 1|y}^2-p_{0, 1|y}p_{1, 0|y}^2-p_{0, 1|y}^2p_{1, 0|y}+p_{0, 0|y}p_{1, 1|y}^2\\
      &=p_{0, 0|y}p_{1, 1|y}-p_{0, 1|y}p_{1, 0|y} = A*D-B*C.
    \end{split}
\end{equation}
Combining (\ref{Eq: std's}) and (\ref{Eq: cov}), we have
\begin{equation}
    \label{Eq:combined}
    \begin{split}
    &\quad \frac{Cov(X_1, X_2|Y=y)}{\sigma_{X_1|Y=y}\sigma_{X_2|Y=y}}\\ 
    & =\frac{A*D-B*C}{\sqrt{(C+D)(A+B)(A+C)(B+D)}}
    \end{split}
\end{equation}
Next, we expand the terms of $\mathcal{H}(X_1|Y=y)+\mathcal{H}(X_2|Y=y)-\mathcal{H}(X_1, X_2|Y=y)$ as follows:

\begin{equation}
\begin{split}
    &\quad \mathcal{H}(X_1|Y=y)+\mathcal{H}(X_2|Y=y)-\mathcal{H}(X_1, X_2|Y=y)\\
    &= p_{0,0|y}\log \frac{p_{0,0|y}}{(p_{0,0|y}+p_{0,1|y})(p_{0,0|y}+p_{1,0|y})}\\
    &\quad + p_{0,1|y}\log \frac{p_{0,1|y}}{(p_{0,0|y}+p_{0,1|y})(p_{0,1|y}+p_{1,1|y})}\\
    &\quad + p_{1,0|y}\log \frac{p_{1,0|y}}{(p_{0,0|y}+p_{1,0|y})(p_{1,0|y}+p_{1,1|y})}\\
    &\quad + p_{1,1|y}\log \frac{p_{1,1|y}}{(p_{0,1|y}+p_{1,1|y})(p_{1,0|y}+p_{1,1|y})}\\
    &= A\log \frac{A}{(A+B)(A+C)} + B\log \frac{B}{(A+B)(B+D)}\\
    &\quad + C\log \frac{C}{(A+C)(C+D)} + D\log \frac{D}{(B+D)(C+D)}
\end{split}
\end{equation}
Using the concavity of logarithm, we further have
\begin{equation}\label{Eq: incremental entropy arrange terms}
\begin{split}
&\quad \mathcal{H}(X_1|Y=y)+\mathcal{H}(X_2|Y=y)-\mathcal{H}(X_1, X_2|Y=y)\\
&\leq \log \bigg[ \frac{A^2}{(A+B)(A+C)} + \frac{B^2}{(A+B)(B+D)}\\
& \quad \quad \quad + \frac{C^2}{(A+C)(C+D)} + \frac{D^2}{(B+D)(C+D)} \bigg]\\
&\leq \log \bigg[2\Big|\frac{Cov(X_1, X_2|Y=y)}{\sigma_{X_1|Y=y}\sigma_{X_2|Y=y}}\Big|^2\\
&\quad +\frac{(A*B*C +A*B*D +A*C*D+B*C*D)}{(C+D)(A+B)(A+C)(B+D)}\bigg]\\
&\leq \log (2\rho^2+1),
\end{split}
\end{equation}
where the last two inequalities follow (\ref{Eq:combined}) and that
\begin{equation}
    \Big|\frac{Cov(X_1, X_2|Y=y)}{\sigma_{X_1|Y=y}\sigma_{X_2|Y=y}}\Big|\leq \rho.
\end{equation}
The above inequality holds for all $y\in \{0, 1\}$, therefore,
\begin{equation}\label{Eq: entropy final final}
\begin{split}
&\quad \mathcal{H}(X_1|Y)+\mathcal{H}(X_2|Y)-\mathcal{H}(X_1, X_2|Y)\\
&=1/d \leq \log (2\rho^2+1).
\end{split}
\end{equation}

By inserting (\ref{Eq: entropy final final}) into (\ref{Eq: proposition}), we finish the proof of Proposition 1.
\end{proof}

\section{Datasets Availability}
\label{datasets}
\begin{itemize}
    \item \textbf{Employee} https://www.kaggle.com/c/amazon-employee-access-challenge/
    \item \textbf{Bank} https://www.kaggle.com/brijbhushannanda1979/bank-data
    \item \textbf{Adult} https://archive.ics.uci.edu/ml/datasets/adult
    \item \textbf{Credit} https://www.kaggle.com/c/GiveMeSomeCredit/data
    \item \textbf{Criteo} http://labs.criteo.com/2014/02/download-kaggle-display-advertising-challenge-dataset/
\end{itemize}

\textbf{Business1} and \textbf{Business2} are constructed by randomly sampling large-scale search logs.
Due to the privacy issue, they are not publicly available currently.

\section{Implementation and HPO Details}
\label{Implementation details of baselines}
For FIVES and all baseline methods, we empirically set batch size as 128 for small datasets (Employee, Bank, Adult and Credit), and 1024 for large datasets (Criteo, Business1 and Business2). The learning rate of baseline methods is set to be 5e-3. To overcome the overfitting issue, model parameters are regularized by $L^2$ regularization with the strength of 1e-4 and the dropout rate is set to 0.3. We apply grid search for hyperparameter optimization (HPO).
After the HPO procedure, we use the optimal configuration to train and evaluate each method for 10 times, alleviating the impact of randomness.
For better parallelism and economic usage of computational resources, we conduct all our experiments on 
Alibaba Cloud\footnote{https://www.alibabacloud.com/product/machine-learning}.
Other implementation details for each method are described as below:

\textbf{FIVES}. The tunable hyperparameters contain the highest order number of interactive features $K \in \{2,3,4\}$, the learning rate $\alpha_1 = $5e-3 and $\alpha_2 \in $\{5e-3, 5e-4\}, the embedding dimension of node representation $\in \{8,16\}$. The hidden dimension of GNN is set to be the same as the embedding dimension. 

\textbf{DNN}. It is implemented by ourselves. The tunable hyperparameters contain the hidden dimension of fully connected layers $\in \{[512,256], [256,128], [128,64]\}$ and the embedding dimension of node representation $\in \{8,16\}$.

\textbf{FM}. The cloud platform we used has provided some frequently used machine learning algorithms including FM.
The tunable hyperparameters include learning rate $\in\{0.005, 0.01\}$ and coefficients of regularization $\in\{[0.001,0.001,0.001],[0.01,0.01,0.01]\}$.

\textbf{Wide\&Deep}. It is implemented by ourselves according to \cite{cheng2016wide}. The tunable hyperparameters contain the embedding dimension of node representation $\in \{8,16\}$ and the hidden dimension of fully connected layers in the deep component $\in \{[512,256], [256,128], [128,64]\}$.

\textbf{AutoInt}. It is reproduced by using the source code\footnote{https://github.com/DeepGraphLearning/RecommenderSystems} published by ~\cite{song2019autoint}. The tunable hyperparameters contain the number of blocks $\in [2,3,4]$ and the number of attention heads $\in [2,4]$. The hidden dimension of interacting layers is set to 32 as suggested by the original paper and the embedding dimension of node representation $\in \{8,16\}$.

\textbf{Fi-GNN}. It is reproduced by ourselves via TensorFlow according to the source code\footnote{https://github.com/CRIPAC-DIG/Fi\_GNN} published by ~\cite{li2019fi}. The tunable hyperparameters contain the highest order number of interactive features $K \in \{2,3,4\}$, and the embedding dimension of node representation $\in \{8,16\}$.

\textbf{AutoFIS}. It is reproduced by using the source code\footnote{https://github.com/zhuchenxv/AutoFIS} published by ~\cite{liu2020autofis}. The tunable hyperparameters contain the embedding dimension of node representation $\in \{8,16\}$.

\textbf{AutoCross}. We implement a special LR that either updates all the trainable parameters or updates only the parameters of newly added features. Then we implement a scheduler to trigger training and evaluation routines of the LR over different feature spaces. This fails to exploit the ``reuse'' trick proposed in~\cite{luo2019autocross}, but identically expresses their search strategy.

\textbf{LR}.  We use the LR provided by the cloud platform, which is implemented based on parameter-server architecture. We set the $L^1$ regularization strength as 1.0 and $L^2$ regularization as 0. The maximum iteration is 100 and the toleration is 1e-6. 

\end{document}